%% file: main.tex
\documentclass[10pt,twocolumn,letterpaper]{article}

\usepackage{cvpr}              % To produce the CAMERA-READY version
\input{preamble}

\definecolor{cvprblue}{rgb}{0.21,0.49,0.74}
\usepackage[pagebackref,breaklinks,colorlinks,citecolor=cvprblue]{hyperref}

\def\paperID{37} % *** Enter the Paper ID here
\def\confName{CVPR}
\def\confYear{2024}

\title{AnimalFormer: Multimodal Vision Framework for Behavior-based Precision Livestock Farming}

\author{Ahmed Qazi \quad Taha Razzaq \quad Asim Iqbal\thanks{Corresponding author.}\\
Tibbling Technologies\\
{\tt\small asim@tibbtech.com}}
\begin{document}
\maketitle
\input{sec/0_abstract}    
\input{sec/1_intro}
% \input{sec/2_formatting}
% \input{sec/3_finalcopy}
{
    \small
    \bibliographystyle{ieeenat_fullname}
    \bibliography{main}
}


\end{document}

%% file: preamble.tex
\usepackage[dvipsnames]{xcolor}

%% file: sec/0_abstract.tex
\begin{abstract}
% This paper introduces a comprehensive framework designed to process and analyze video data for insights into sheep behavior, leveraging state-of-the-art deep learning techniques. Utilizing a primary dataset capturing five distinct types of sheep activities—grazing, running, sitting, standing, and walking—the framework employs the GroundingDINO model to accurately generate bounding boxes around the subjects and the HQ-SAM model for segmenting each sheep within these bounding boxes. Additionally, we incorporate the VitPose model for precise keypoint estimation, enabling an in-depth analysis of posture and movement patterns. This approach transforms raw video input into actionable insights, such as activity patterns, behavioral anomalies, interaction dynamics, and detailed posture analysis, without any invasive tagging or direct animal contact. Expanding domain knowledge for virtual fencing applications, our framework also significantly impacts non-invasive sheep activity detection, counting, health status monitoring, and posture analysis. Its ability to process videos at various resolutions and viewing angles ensures robustness and adaptability, offering a powerful tool for modern farming activities and animal behavior research.

We introduce a multimodal vision framework for precision livestock farming, harnessing the power of GroundingDINO, HQSAM, and ViTPose models. This integrated suite enables comprehensive behavioral analytics from video data without invasive animal tagging. GroundingDINO generates accurate bounding boxes around livestock, while HQSAM segments individual animals within these boxes. ViTPose estimates key body points, facilitating posture and movement analysis. Demonstrated on a sheep dataset with grazing, running, sitting, standing, and walking activities, our framework extracts invaluable insights: activity and grazing patterns, interaction dynamics, and detailed postural evaluations. Applicable across species and video resolutions, this framework revolutionizes non-invasive livestock monitoring for activity detection, counting, health assessments, and posture analyses. It empowers data-driven farm management, optimizing animal welfare and productivity through AI-powered behavioral understanding.
\end{abstract}

 % anomaly detection

%% file: sec/1_intro.tex
\section{Introduction}
\label{sec:intro}
The fusion of computer vision and deep learning has catalyzed a paradigm shift in non-contact animal monitoring systems. These advanced systems, now a cornerstone in modern agriculture, are pivotal for animal behavior quantification and early disease detection~\cite{chicken_pose}. As farming moves towards precision livestock farming, the role of such technologies in ensuring animal health and welfare has become indispensable. Within this domain, pose estimation~\cite{pose_estimation_mice, Toshev_2014, he2018mask} and instance segmentation~\cite{segmentation_analytics, he2018mask} emerge as two pillars foundational to behavioral analysis, driving efforts towards creating robust and efficient models. Pose estimation, a domain thoroughly explored, benefits from extensive datasets ~\cite{ap10k, apt36k} that provide a wealth of annotated information across diverse animal species. Leveraging such resources, generalizable pose estimation models~\cite{vitpose, superanimal} achieve remarkable results, enhancing our understanding of animal behavior across the animal kingdom. While pose estimation offers in-depth behavioral insights, instance segmentation opens doors to analytics that transcend posture analysis~\cite{segmentation_analytics}. However, the isolated application of these techniques only provides a fraction of the potential insight.

Current approaches for evaluating the behavioral states of farm animals, such as direct observation or physiological measurements, are often disruptive, subjective, and impractical for large-scale application \cite{cooper2020qualitative}. There's a lack of scientifically validated metrics to accurately gauge farm animal well-being using these methods \cite{neethirajan2021measuring}. Standards in animal welfare predominantly highlight negative indicators, like harm and stress, rather than positive ones \cite{viscardi2017development, mogil2020development}. Yet, recognizing and promoting positive states are crucial not only for the animals' health but also for enhancing farm productivity and product quality. Non-invasive monitoring offers a promising solution to these challenges. By leveraging technologies that minimize stress and bias, we can gather reliable data on animal welfare, potentially improving both the lives of livestock and the outputs they yield \cite{animal_welfare_framework, mota2020effects}. The importance of such advancements cannot be overstated. With the global population projected to reach nearly 10 billion by 2050 \cite{Lutz2010DimensionsOG}, the demand for sustainable and efficient farming practices is at an all-time high.

\begin{figure*}[!htb]
    \centering
    % \hspace{0mm}
    \includegraphics[scale=0.66]{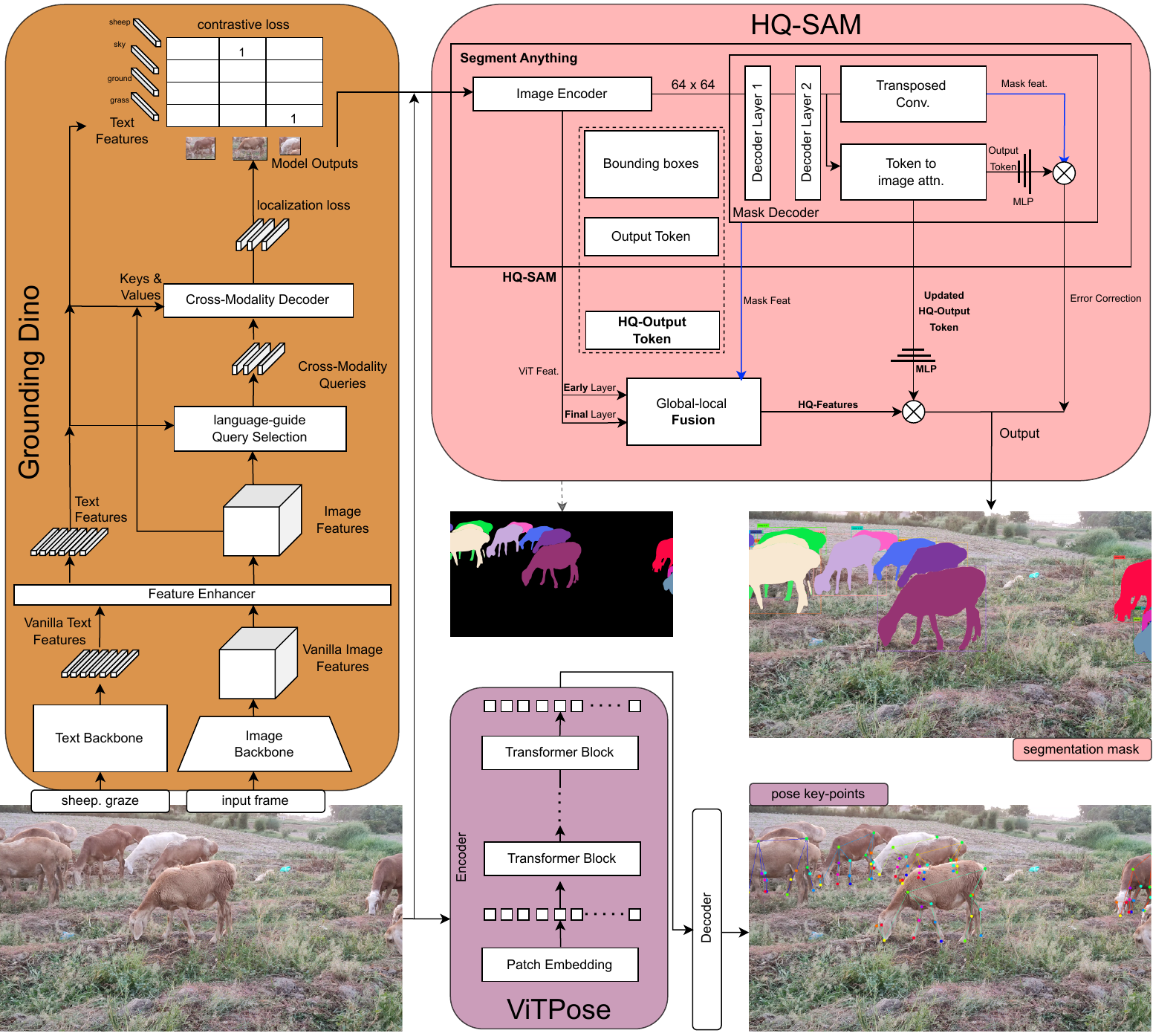}
    % \vspace{-0.2cm}
    \caption{Our integrated analysis framework, designed for comprehensive behavioral understanding of sheep in a dataset. The framework combines ViTPose and Grounding DINO for pose estimation and contextual understanding respectively, with the high-quality instance segmentation capabilities of HQ-SAM. By fusing these components, our pipeline provides precise keypoints and segmentation masks, essential for in-depth ethological studies. This block diagram provides a clear overview of the data flow and processing steps within our end-to-end solution.}
    \label{fig:fig1}
    \vspace{-0.3cm}
\end{figure*}

We introduce a revolutionary non-contact, multimodal AI framework, AnimalFormer, that synergizes state-of-the-art models for a holistic analysis of livestock behavior. Utilizing an open-source sheep dataset \cite{kelly2024videodataset}, we demonstrate comprehensive behavioral analytics across various activities such as grazing, running, and sitting/resting. The combination of ViTPose~\cite{vitpose}, GroundingDINO~\cite{grounding_dino}, and HQ-SAM~\cite{hq_sam} allows for nuanced pose estimation and precise segmentation, capturing the granularity of animal movements and interactions. AnimalFormer provides a scalable solution that can be adapted across different livestock species and varying farm sizes, offering insights that are critical for optimizing feed efficiency, monitoring health, and potentially improving breeding programs. This framework transcends the capabilities of existing methodologies, offering a novel approach to farm management that is both advanced and operationally efficient. By harnessing raw video data, our framework as shown in {\hyperref[fig:fig1]{\textbf{Figure 1}}}, facilitates precision livestock farming and proactive health monitoring without the need for invasive tagging or interaction. The insights gleaned are not only pivotal for individual animal welfare but also hold the promise of transforming farming practices on a global scale. The non-invasive nature of our technology ensures minimal stress to animals, leading to more accurate behavioral observations and, consequently, a deeper understanding of animal well-being. 

% Moreover, the ability to detect and analyze subtle behavioral changes in real-time opens new avenues for early disease detection and prevention, potentially saving significant costs and reducing animal suffering. Furthermore, the implications for animal welfare are profound. By enabling a deeper understanding of animal behavior in a non-invasive manner, farmers and researchers can make informed decisions that prioritize the well-being of their livestock. This, in turn, can lead to improvements in farm productivity and sustainability, as healthy and well-managed animals are known to perform better.

To the best of our knowledge, such a comprehensive, multimodal AI framework using latest state-of-the-art Vision Transformers (ViTs) \cite{dosovitskiy2020image} models has not been previously devised specifically for the agriculture and farming community. By bridging this gap, we not only contribute to the advancement of precision livestock farming but also set a new benchmark for animal welfare and farm management practices. The potential for scaling this framework to incorporate additional behavioral analytics and apply it to a wider range of animal species further emphasizes its significance and the broad impact it could have on the future of farming and agricultural industry.

\section{Related work}
The advancement in non-contact animal monitoring systems, particularly through pose estimation, has significantly improved our understanding of animal behavior. Large-scale open-sourced datasets, combined with deep learning techniques, have fostered the development of efficient pose estimation models \cite{superanimal}. These models enable detailed behavior analysis across various species, offering insights into their natural patterns and interactions. Notably, mice have been a focal point in behavior analytics, leading to models that capture their diverse behaviors \cite{pose_estimation_mice, mouse_pose}. In livestock farming, pose estimation has proven invaluable. It facilitates behavior analytics for cattle, aiding in breeding reference, early detection of lameness \cite{lameness}, and monitoring overall health \cite{lameness_gait}. This approach, free from human intervention, not only ensures comprehensive behavior analytics but also reduces the risk of injury to animals. However, pose estimation alone may not capture the full spectrum of animal interactions, especially in scenarios involving occlusion or requiring the analysis of intricate behaviors such as eating patterns. This limitation stems from the reliance of pose estimation models on keypoints, which might not effectively capture changes across consecutive frames. To address these gaps, instance segmentation emerges as a powerful tool, enabling the detailed tracking of animal interactions and behaviors \cite{segmentation_analytics}. Coupled with multiple-object detection, this approach enhances the tracking and trajectory analysis of animals within video footage.

Recognizing the strengths and limitations of both pose estimation and instance segmentation, we propose an end-to-end multimodal framework that integrates these methodologies. This framework aims to provide comprehensive behavioral analytics, encompassing a broader range of activities and interactions. Through this synthesis, we seek to offer a more holistic view of animal behavior, bridging the gap between individual and collective behaviors.

\begin{figure*}[!htb]
    % \centering
    \hspace{1.5mm}
    \includegraphics[scale=0.37]{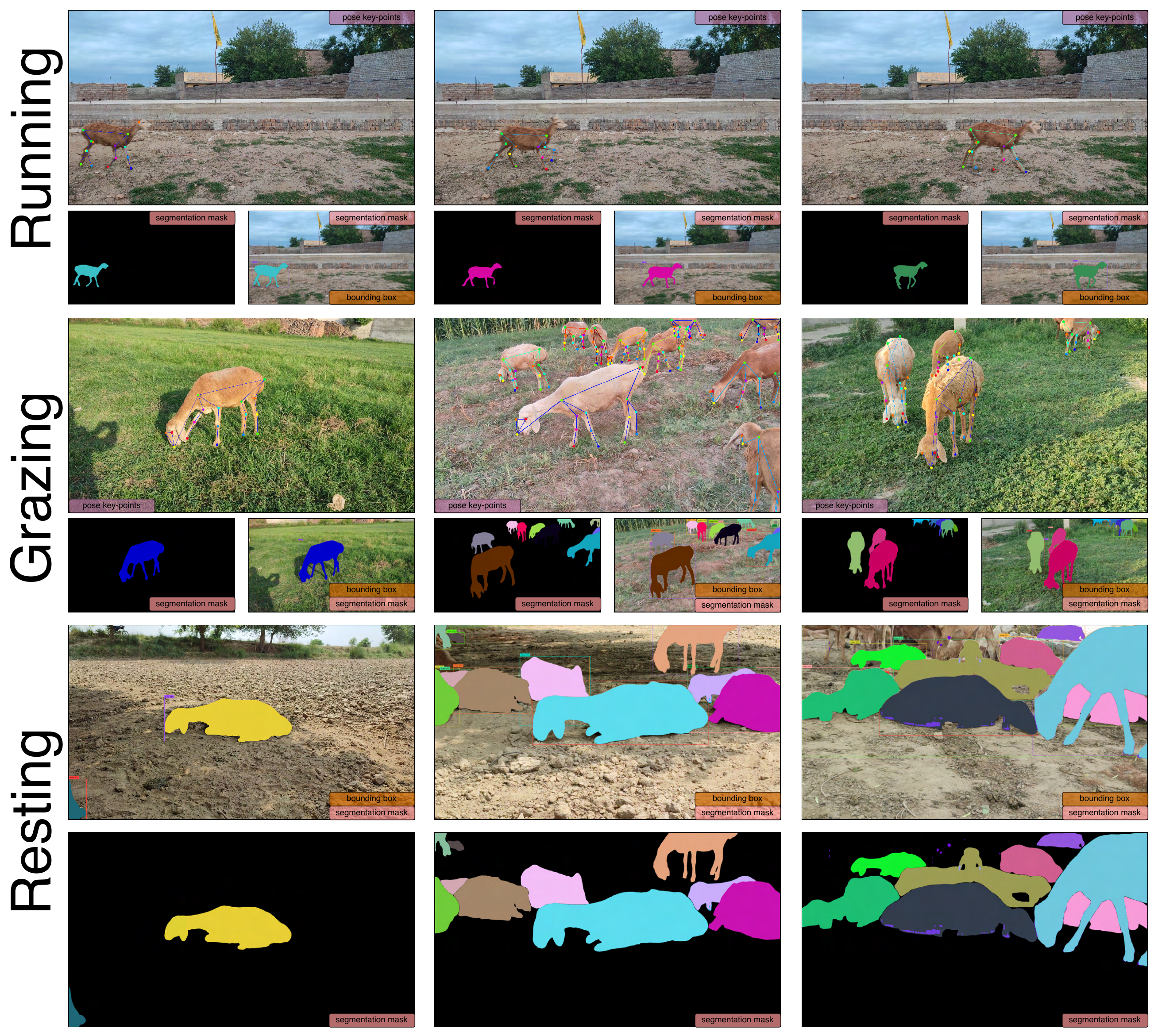}
    % \vspace{-0.2cm}
    \caption{Qualitative outputs of our framework depicting various behaviors of sheep. Top row: Running frames with keypoints and bounding boxes illustrating movement dynamics. Middle row: Grazing frames showcasing sheep engaged in feeding with extracted poses and segmentation masks highlighting the focus areas. Bottom row: Resting frames with segmentation masks delineating individual sheep in a state of repose. Each behavior is analyzed through a combination of visual features extracted from images.}
    \label{fig:fig2}
    \vspace{-0.3cm}
\end{figure*}

\section{Methods}
\subsection{Dataset}
The dataset \cite{kelly2024videodataset} used in this paper was curated for sheep behavior analysis. It constitutes of an extensive collection of videos capturing five distinct sheep activities: \textit{grazing}, \textit{running}, \textit{sitting}/resting, \textit{standing}, and \textit{walking}. The videos were recorded using high-resolution digital cameras from varying angles and positions, and the entire dataset comprises of $417$ videos. Although, the duration of each video varies, the total footage is $59$ minutes long, totaling to $149,327$ frames. For our analysis, we only consider the videos pertaining to grazing, running and sitting/resting activities. Furthermore, to enhance computational efficiency, we downsampled the dataset by selecting every $10^{th}$ frame. This downsampling process reduces the computational load while preserving essential information for analysis, ensuring the dataset remains suitable for developing and testing computer vision algorithms aimed at non-invasive monitoring and analysis of sheep behavior. 

% This diverse compilation aims to enhance the scope of non-contact virtual fencing technologies and contributes significantly to animal behavior analysis. The dataset's variety in terms of resolution, viewing angles, and sheep activities provides a robust foundation for developing and testing advanced computer vision techniques aimed at non-invasive monitoring and analysis of sheep behavior. An important aspect of this dataset is its contribution towards enabling the development of models for accurate sheep activity detection, a pivotal step towards efficient farm management and animal welfare. 
The activities captured in the dataset are reflective of typical behaviors exhibited by sheep in a natural setting, making it a valuable resource for training deep learning models to recognize and classify these activities accurately. The dataset's format, consisting of high-resolution videos and their corresponding frames (with a notable omission of frames containing humans for privacy), allows for a wide array of computer vision applications, including but not limited to, activity recognition and behavior analysis. 

\subsection{AnimalFormer}
In AnimalFormer, we integrate ViTPose \cite{vitpose}, Grounding DINO \cite{grounding_dino}, and HQ-SAM \cite{hq_sam} to extract keypoints and perform instance segmentation. These elements collectively contribute insights into the sheep dataset. The comprehensive framework of AnimalFormer is shown in {\hyperref[fig:fig1]{\textbf{Figure 1}}}.

\subsubsection{Pose estimation}
We utilize ViTPose \cite{vitpose}, the current state-of-the-art pose estimation model, to extract keypoints for each video. ViTPose provides a simple but efficient ViT baseline for pose estimation in animals as well as human datasets. It uses plain and hierarchical vision transformers, as the backbone model which is trained on ImageNet-1K \cite{deng2009imagenet} with masked image modelling (MIM) \cite{mim} pre-training. ViT is coupled with light weight decoders which bypass the use of complex mechanisms including skip-connections and cross-attentions and instead comprise of simple deconvolution and prediction layers. 

An input image $X \in \mathcal{R}^{H \times W \times 3}$, is first converted into tokens via the patch embedding layer i.e $F \in \mathcal{R}^{\frac{H}{d} \times \frac{W}{d} \times C}$ where $d$ is the downsampling ratio for each patch embedding layer and $C$ is the number of channels in the input image. The embedded tokens are then fed to the transformer layer which converts them to feature i.e $F_{out} \in \mathcal{R}^{\frac{H}{d} \times \frac{W}{d} \times C}$. Each transformer layer consists of multi-head self attention layer (MHSA) along with a feed-forward network (FFN) i.e.
\begin{equation}
    F'_{i+1} = F_i + \text{MHSA}(\text{LN}(F_i))
\end{equation}
\begin{equation}
    F_{i+1} = F'_{i+1} + \text{FFN}(\text{LN}(F'_{i+1}))
\end{equation}
where $i$ is the output of the $i$th transformer layer and the initial feature ($F_0$) is the output of the patch embedding layer. LN is the normalization layer.

The features extracted by the ViT are fed to the decoders which upsamples them twice before converting them to keypoints $K \in \mathcal{R}^{\frac{H}{4} \times \frac{W}{4} \times N_k}$ where $N_k$ is the number of keypoints to be estimated, which is set to $17$ for animal datasets.
% The core attention mechanism is described as follows:

% \begin{equation}
% \text{Attention}(Q, K, V) = \text{softmax}\left(\frac{QK^T}{\sqrt{d_k}}\right)V,
% \end{equation}

% where \(Q\), \(K\), and \(V\) denote the queries, keys, and values matrices derived from the input, and \(d_k\) is the dimensionality of the keys. This mechanism facilitates the identification of relevant visual regions corresponding to textual queries.
% Training involves minimizing the cross-entropy loss between the student and teacher networks' output distributions, fostering "local-to-global" correspondences to enhance learned representations.

\subsubsection{Animal detection}
For our animal detection module, GroundingDINO \cite{grounding_dino} is utilized, building on the DINO framework's self-supervised learning strengths. DINO, underpinned by the distilled knowledge self-attention \cite{caron2021emerging, hinton2015distilling} mechanism from ViTs, empowers our model to recognize textual descriptions within visual inputs efficiently. The DINO methodology incorporates self-supervised learning strategies \cite{he2020momentum, chen2020simple, grill2020bootstrap}, utilizing an exponential moving average of the student model's parameters to update the teacher model, thus preventing collapse without a direct contrastive loss \cite{chen2020simple}. 

The distilled knowledge within the student network is refined through temperature-scaled softmax functions, as expressed below:

\begin{equation}
P_s(x)^{i} = \frac{\exp(g_{\theta_s}(x)^{i}/\tau_s)}{\sum_{q=1}^{Q} \exp(g_{\theta_s}(x)^{q}/\tau_s)},
\end{equation}

where \( P_s(x)^{i} \) denotes the probability predicted by the student network for the \( i^{th} \) class, \( g_{\theta_s}(x)^{i} \) is the logit or the raw output from the student network for the \( i^{th} \) class before applying softmax, \(\tau_s\) represents the temperature parameter influencing the softmax function's sharpness and the sum in the denominator runs over all \( Q \) possible classes, ensuring that the probabilities sum to 1. A similar equation pertains to the teacher network \(P_t\) with temperature \(\tau_t\). 

Grounding DINO advances this architecture by precisely linking textual phrases with their corresponding objects in images, for instance, identifying a cat and a table from an input image and associating them with their respective labels extracted from the input text. This enhanced capability is achieved through an image backbone for extracting visual features, a text backbone for text feature extraction, a feature enhancer for cross-modality fusion, and a cross-modality decoder for refining object boxes.

\subsubsection{Animal segmentation}
Our segmentation utilizes HQ-SAM \cite{hq_sam}, an adept extension of SAM \cite{kirillov2023segment}, for its proficiency in generating high-quality masks. The model introduces the \textit{HQ-Output Token}, improving dynamic convolution through a dedicated three-layer MLP (\textit{MLP\textsubscript{dyn}}) and an attention mechanism (\textit{Attn}) that refines the token with input features (\textit{F\textsubscript{in}}):
\begin{equation}
\begin{split}
\text{HQ-Mask} = \text{MLP}_{\text{dyn}}(\text{HQ-Output Token}) \odot \text{F}_{\text{HQ-F}},
\end{split}
\end{equation}
\begin{equation}
\begin{split}
\text{HQ-Output Token}_{\text{upd}} = \text{Attn}(\text{HQ-Output Token}, \text{F}_{\text{in}}),
\end{split}
\end{equation}

with the enhanced output feature (\textit{F\textsubscript{out}}) attained by element-wise multiplication of the softmax-weighted input feature map:
\begin{equation}
\begin{split}
\text{F}_{\text{out}} = \text{softmax}\left(\text{W}_f \cdot \text{F}_{\text{in}}\right) \odot \text{F}_{\text{in}},
\end{split}
\end{equation}
Global-local fusion combines contextual and boundary detail features across network stages, culminating in a comprehensive feature map (\textit{F\textsubscript{HQ-Fs}}):
\begin{equation}
\begin{split}
\text{F}_{\text{HQ-Fs}} = \text{Conv}\left(\text{F}_{\text{early}} \oplus \text{F}_{\text{final}} \oplus \text{F}_{\text{mask}}\right),
\end{split}
\end{equation}
HQ-SAM thus upholds SAM’s zero-shot segmentation efficacy with substantial quality improvements for practical, high-fidelity segmentation tasks.

\begin{figure*}[!htb]
    \centering
    \hspace{-1mm}
    \includegraphics[scale=0.56]{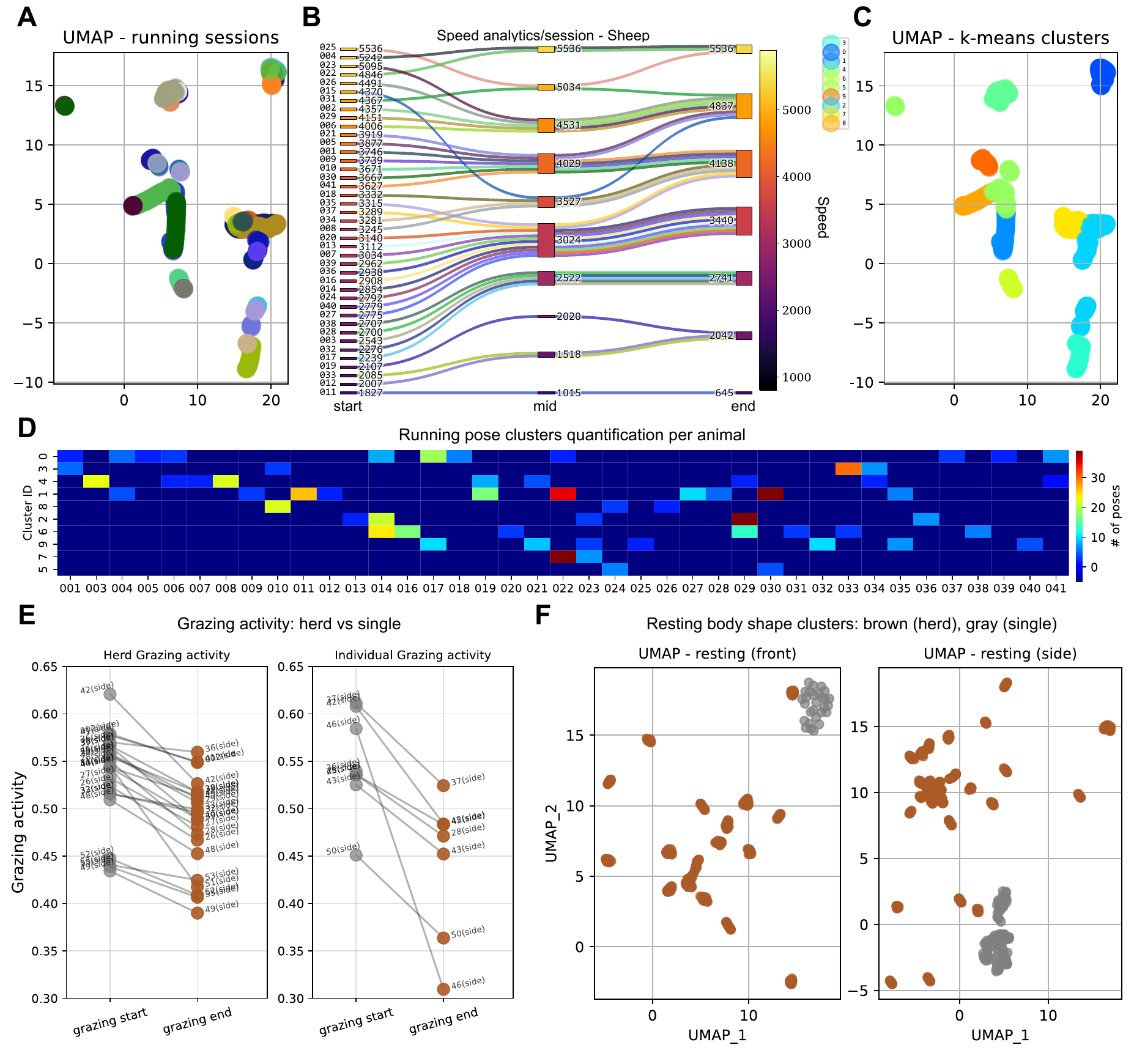}
    % \vspace{-0.2cm}
    \caption{Behavior Analytics. \textbf{A.} UMAP representation of the unique gait patterns of the sheep extracted from their running videos. \textbf{B.} Speed profile of the animal extracted at the commencement, midpoint, and conclusion of their running videos. \textbf{C.} The unique clusters of existing sheep gait patterns. \textbf{D.} Spread of different patterns within a single animal across different clusters. \textbf{E.} The grazing activity of the sheep in herds vs single. \textbf{F.} UMAP representation of the unique resting pattern of the animals in herd vs single.} 
    \label{fig:fig3}
    \vspace{-0.3cm}
\end{figure*}

\subsection{Behavior analytics}
\subsubsection{Running videos}
\noindent \textbf{Gait.} In order to quantify the overall gait pattern of an animal, we look at the varying poses it exhibited during running. A set of $17$ keypoints is extracted from each frame of an animal's running video as shown in {\hyperref[fig:fig2]{\textbf{Figure 2}}}. The keypoints are converted to meaningful embedding by applying Uniform Manifold Approximation and Projection (UMAP) \cite{umap}, which represent the different poses demonstrated by an individual sheep during the course of its running. For UMAP, we set \texttt{n\_neighbors=20} to encapsulate all the poses emphasized by the running of an animal and a two-dimensional embedding space (\texttt{n\_components=2}) is used for visualization. To quantify the unique poses present across the videos and extract similar gait patterns between different sheeps, we cluster the embeddings via K-Means Clustering \cite{kmeans}. We create $10$ unique clusters for all the poses demonstrated by the sheep's in the videos. In order to extract meaningful insights about the sheep behavior we assess how gait patterns affect speed, by conducting a speed analysis of the running videos. We opt for UMAP as part of our pipeline due to its ability to preserve local and global structures within the data along with being highly efficient in processing large volumes \cite{umap}. 
\\
\noindent \textbf{Speed.} We assess the speed of individual subjects based on their observable movement across video frames ({\hyperref[fig:fig2]{\textbf{Figure 2}}}) in running videos. Given a set of masks corresponding to each subject's position in sequential frames, we commence by identifying the largest mask within each frame, which we associate with the primary subject of interest. This is predicated on the notion that the primary subject occupies the largest area within the field of view. To ascertain the centroid of the largest mask, we calculate the geometric center of its binary mask area, which serves as a proxy for the subject's position. The temporal evolution of the centroid between consecutive frames allows us to deduce the raw speed of the subject. This is achieved by measuring the displacement of the centroid over time, considering the frame rate of the video and the number of frames skipped between measurements to enhance accuracy. However, raw speed alone does not provide a complete picture due to potential variations in the subject's distance from the camera, which can skew perceived speed. To mitigate this, we normalize the speeds by the areas of the masks. The normalization process involves adjusting the speeds based on a reference area value derived from the mean of all mask areas. The intent is to correct for perspective distortion, wherein objects that are closer to the camera appear to be moving faster than those further away. By normalizing the speeds, we aim to achieve a more accurate representation of the actual speed of the subjects. Our methodology yields a set of normalized speeds for each subject, corresponding to their movement across the frames. This data along with the gait patterns forms the basis for subsequent analyses of the sheep.

\subsubsection{Grazing videos}
To perform analysis of the grazing behavior among sheep, we begin with segmenting individual sheep across video frames as shown in ({\hyperref[fig:fig2]{\textbf{Figure 2}}}). After accurate identification and segmentation, we determine the keypoint representative of the sheep’s nose. This keypoint serves as the focal point around which we establish an approximate grazing area. We further refine this selected grazing patch by subtracting the sheep's segmentation masks to exclude non-grazing elements such as sheep's head, ensuring that only relevant grazing area is measured. Within this isolated region, we quantify the green signal intensity. The measurement of the green signal intensity across consecutive frames provides an approximation of the grazing activity which was indicative of the amount of grass intake. This allowed us to make inferential analyses about the grazing patterns among the sheep.

\subsubsection{Resting videos}
On sheep resting behavior, we employed an image-based analysis to differentiate between individual and herd resting states from front and side view ({\hyperref[fig:fig2]{\textbf{Figure 2}}}). The methodology involved segmenting video footage into frames, using bounding boxes to identify and isolate sheep within each frame. We extracted masks for each sheep identified by the bounding boxes, categorizing the data into four distinct groups - \textit{front view herd}, \textit{side view herd}, \textit{front view single}, and \textit{side view single}. The masks were resized to a standard dimensions for consistency across the dataset. The processed data was then analyzed using UMAP for dimensionality reduction, allowing us to visualize and cluster different resting states. UMAP's configurations were set to \texttt{n\_neighbors=50} to account for the overarching grouping tendencies of sheep, and \texttt{min\_dist=0.01} to achieve distinct clusters indicative of individual and group resting states. A two-dimensional space projection with \texttt{n\_components=2} was chosen for ease of visualization and analysis, with the Euclidean metric due to its natural fit for the data's geometric properties. To quantify the unique resting poses, we cluster the UMAP embeddings via K-Means Clustering. We create $10$ unique clusters for all the poses demonstrated by the sheep's in the sitting videos. This approach facilitated the identification of patterns in sheep resting behavior, such as preferences for resting alone or within a group.

\section{Results \& Discussion}

\subsection{Indirect relationship between animals' gait diversity and running speed}
Since sheep are prey animals with extremely heightened senses of flight and fright, they tend to run away from humans and other animals \cite{Dwyer_2004}. Running is a common behavioral habit observed in sheep which can tell us a great deal about their gaits and speed. Using the running videos of individual sheep and their corresponding UMAP embeddings, we extract the different poses exhibited by the sheep, as shown in {\hyperref[fig:fig3]{\textbf{Figure 3 (A)}}}. Each unique color in {\hyperref[fig:fig3]{\textbf{Figure 3 (A)}}} represents an individual animal's poses. We observe that although for majority animals, the poses exhibited lie closer to each other however in some instances they are spread out. To validate this observation, we cluster the animal poses into $10$ unique groups as shown in {\hyperref[fig:fig3]{\textbf{Figure 3 (C)}}}. Although the running patterns of a sheep are independent of other subjects, we observe that some sheep share gait patterns as they belong to the same cluster. {\hyperref[fig:fig3]{\textbf{Figure 3 (D)}}} shows the different poses encompassed within an individual animals running pattern along with the distribution of poses in multiple clusters. Even though different sheep have overlapping gait patterns, however {\hyperref[fig:fig3]{\textbf{Figure 3 (D)}}} clearly shows that an individual sheep's gait pattern is consistent within itself and is largely found within a single cluster. Animal IDs \textit{025} and \textit{011}, for example, have a high number of poses, which are found in a single cluster i.e cluster ID 9 and 1 respectively. We observe a similar trend in case of a dynamic gait pattern spanning multiple clusters and observe that majority of the poses are present in a specific cluster. Animal ID \textit{014}, for example, has poses spanning clusters 0, 2 and 6 however majority of its poses are present in cluster 6 while a few are distributed among 0 and 2. A similar trend is observed for other animals with higher number of unique poses, including animal IDs \textit{029} and \textit{033}, for which majority of their poses are present in cluster 2 and 0 respectively. 

Although, pose estimation provides insights regarding the gait patterns, however in order to quantify the animal's health, we further provide speed analysis. {\hyperref[fig:fig3]{\textbf{Figure 3 (B)}}} illustrates the speed trends of individual sheep at various intervals - the commencement, midpoint, and conclusion of their running sessions. The figure presents a compelling narrative; sheep that display a more limited range of dynamic gait patterns tend to achieve higher speeds. Animal ID \textit{025}, for example, encompasses few poses within its running pattern, however it is the fastest among the lot. On the other hand, animal ID \textit{014}, is among the slower running sheep despite having a higher number of poses. This phenomenon suggests a possible trade-off between speed and the variety of movements within the gait. In essence, sheep that adhere to a consistent gait pattern, without significant variations, generally exhibit higher velocities. Conversely, sheep with a broader array of dynamic poses show a tendency toward slower speeds. The observed inverse relationship aligns with the principles of locomotive efficiency, where streamlined, repetitive movements facilitate faster motion \cite{ariani2020repetita}. This insight is not only of academic interest but also carries practical implications. Understanding the gait diversity and corresponding speed could aid in identifying potential health issues, optimizing shepherding practices, and enhancing overall herd management.

\subsection{Enhanced grazing activity in isolation}
Through animal detection and pose estimation, we quantify and compare the grazing patterns of sheep in an isolated setting with when they are flocked together. 
Our results, as shown in {\hyperref[fig:fig3]{\textbf{Figure 3 (E)}}}, reveal compelling evidence that sheep engage in more substantial grazing activity when isolated as compared to when they are part of a herd. In absence of social stimuli and potential threats, the sheep focus more intently on feeding and have a higher grazing activity. This behavior is attributed to the diminished distractions and the lack of need for social interaction or vigilance, which is more prevalent in group settings. In controlled environments, devoid of predators or environmental stressors, animals typically exhibit a reduction in the vigilance behavior that characterizes wild settings. Our observations are in alignment with this principle, indicating that solitary grazing sheep capitalize on the safety and tranquility to maximize their food intake. We observe that when the animals flock together, they tend to consume lesser food, since they pause their grazing for various social interactions, such as establishing social hierarchies or bonding. The quantitative data gleaned from our analysis provides a clear contrast between the two scenarios. Sheep in solitude not only graze more but also show consistent feeding patterns without significant interruptions. This contrasts with herd scenarios where the grazing patterns are interspersed with non-feeding activities. Our study offers valuable insights into the grazing patterns of domestic sheep, which can be instrumental in enhancing farm management practices and optimizing feeding strategies to ensure the well-being and productivity of the animals.

In short, our examination of sheep grazing behaviors yielded an intriguing pattern - in controlled scenarios, sheep displayed an increased grazing activity when alone compared to being in a herd. This increase is hypothesized to be due to the reduced distractions and social obligations that typically interrupt feeding in a group environment. Our data suggests that in such isolated snippets, the absence of broader social dynamics allows sheep to feed more efficiently, leveraging the security and stillness of a controlled setting to focus on grazing. It is worth noting, however, that these findings are based on brief observational periods. A long-term comprehensive study could reveal different grazing dynamics, as sheep might exhibit varying grazing intensities over extended periods \cite{herd_grazing}. Our insights are constrained by the scope of the dataset, which did not capture extended durations or include variables like environmental shifts and detailed social interactions, limiting the breadth of our analysis.

\subsection{Resting pose similarities across animals}
In {\hyperref[fig:fig3]{\textbf{Figure 3 (F)}}}, the plot on the left displays the UMAP scatterplot for the front view of resting sheep, while the plot on the right shows the side view. In each plot, data points represent individual instances of sheep resting postures. The clustering of points in the UMAP space signifies patterns of similarity within the resting postures. A tighter clustering indicates a more consistent and homogeneous behavior pattern among the animals within that group. From these plots, we observe that sheep demonstrate a propensity for maintaining a consistent resting posture when alone, which is reflected in the tightly clustered gray points. This homogeneity in behavior could be attributed to a lack of external stimuli, leading to a uniformity in resting positions. In contrast, the brown clusters corresponding to sheep resting in a herd show a more scattered distribution, suggesting a diversity in resting postures. This variability could arise from social dynamics within the herd or because of spatial constraints. The scattered clusters imply that sheep in a herd may change their resting positions more frequently or adopt different postures compared to when they are alone. The UMAP plots provide a compelling visual summary of the resting behaviors, highlighting the differences between solitary and social resting states in sheep. This insight is crucial for understanding sheep behavior in controlled environments and has implications for animal welfare and the management of livestock.
% The data supports the grazing analytics findings that sheep exhibit more consistent patterns in behavior when alone.
These UMAP visualizations eloquently encapsulate the dichotomy between solitary and social resting states among sheep. The implications of this data are profound, aligning with the grazing behavior findings and reinforcing the notion that sheep display more uniform behaviors when solitary. Such insights are pivotal for the realm of animal husbandry, particularly in the context of enhancing livestock management and welfare practices.
\section{Conclusion}
Our study presents a robust multimodal AI-based vision framework that integrates cutting-edge models for comprehensive behavioral analytics applicable to precision livestock farming. Through the analysis of a sample sheep dataset, we have showcased the utility of our framework in extracting detailed insights into various sheep activities such as running, grazing, and resting. The application of our framework has revealed fascinating observations that are of significant value to the agricultural and farming communities. For instance, the solitary grazing patterns of sheep in a controlled environment without apparent dangers suggest potential adjustments in farm management practices. The insights into their resting patterns further emphasize the impact of social dynamics on animal behavior, which could guide enhancements in livestock care and facility design. Similarly, our insights from sheep running videos validate the finding that repetitive movements result in faster speed \cite{ariani2020repetita}. However, the current dataset did not encompass certain activities, such as milking, and these findings are based on brief observational videos, limiting the scope of our analyses to report their health. Additionally, while we focused on sheep, extending this framework to other species (e.g. cows) could provide a broader understanding of livestock behaviors and welfare across different farming systems.
% that would allow for a more extensive analysis of the animals' health and productivity. Furthermore, 

In future, our goal is to apply this framework to diverse species and environmental conditions.
% to build a more comprehensive animal behavior analysis platform. 
This will enable us to not only continue our exploration of animal activities but also investigate aspects such as dietary habits in relation to the nutritional content of forage. By correlating behavior with nutritional intake, future work could offer guidance on pasture management to optimize feeding efficiency and herd health. We believe AnimalFormer has the potential to revolutionise livestock monitoring by providing non-invasive, in-depth behavioral analysis without any additional hardware (e.g. RFID-based animal tagging). Such a multimodal vision framework is indispensable for advancing precision farming operations, promoting animal welfare, and improving the productivity and sustainability of the agricultural sector.
% As we expand our research, we aim to explore and suggest more insightful data-driven recommendations that will not only enhance the well-being of animals but the effectiveness of farm management practices.